\documentclass{article}
\usepackage{fullpage}

\usepackage[utf8]{inputenc} % allow utf-8 input
\usepackage[T1]{fontenc}    % use 8-bit T1 fonts
\usepackage{url}            % simple URL typesetting
\usepackage{booktabs}       % professional-quality tables
\usepackage{amsfonts}       % blackboard math symbols
\usepackage{nicefrac}       % compact symbols for 1/2, etc.
\usepackage{microtype}      % microtypography
\usepackage{soul}
\usepackage{graphicx}
\usepackage{amsmath}
\usepackage{outlines}
\usepackage{xurl}

\usepackage[colorlinks=true,
citecolor=blue,
filecolor=black,
linkcolor=blue,
urlcolor=blue]{hyperref}

\usepackage{amsmath,amsfonts,bm}

\usepackage{xcolor}
\usepackage{colortbl}

\def\eqref#1{equation~\ref{#1}}
\DeclareMathAlphabet{\mathsfit}{\encodingdefault}{\sfdefault}{m}{sl}
\SetMathAlphabet{\mathsfit}{bold}{\encodingdefault}{\sfdefault}{bx}{n}

\newcommand{\R}{\mathbb{R}}

\usepackage{multirow}
\usepackage{paralist}

\usepackage{multicol}
\usepackage{diagbox}

\usepackage[ruled,noend]{algorithm2e}

\SetCommentSty{mycommfont}

\usepackage{here}

\usepackage{amsmath,amssymb,amsfonts,amsbsy,amsfonts,latexsym}
\usepackage{makecell}
\usepackage{xcolor}
\usepackage{colortbl}

\usepackage{tabularx,colortbl,xcolor}
\usepackage[normalem]{ulem}
\useunder{\uline}{\ul}{}

\usepackage{enumitem}

\usepackage{xparse}% http://ctan.org/pkg/xparse

\SetKwInput{KwInput}{Input}
\SetKwInput{KwRequire}{Require}

\usepackage{longtable}

\NewDocumentCommand{\var}{O{s} m O{}}{%
  \ensuremath{#1_{#2}^{#3}}% add \vphantom{<bizarre sup>}
}
\usepackage{siunitx}

\newcommand{\commentout}[1]{}

\definecolor{light-gray}{gray}{0.80}

\newcommand\fref{Figure~\ref}
\newcommand\tref{Table~\ref}
\newcommand\sref{Section~\ref}

\def\R{{\mathbb R}}

\usepackage{amsthm}

\usepackage{amsthm}

\newtheorem{mytheorem}{Theorem}
\newtheorem{assumption}[mytheorem]{Assumption}
\newtheorem{lemma}[mytheorem]{Lemma}
\newtheorem{remk}[mytheorem]{Remark}

\usepackage{subfig}

\usepackage{pifont}% http://ctan.org/pkg/pifont
\newcommand{\xmark}{\ding{55}}%

\newcommand{\qat}{QAT\xspace}
\newcommand{\ptq}{PTQ\xspace}

\newcommand{\bert}{BERT\xspace}
\newcommand{\gpt}{GPT\xspace}
\newcommand{\ours}{ZeroQuant-HERO\xspace}
\newcommand{\twq}{TWQ\xspace}
\newcommand{\fwq}{FWQ\xspace}
\newcommand{\sq}{SQ\xspace}

\usepackage{chngcntr}
\usepackage{adjustbox}

\begin{document}

% \title{A Practical Post-Training Quantization Framework for Large Scale Generative Transformers}
\title{
\ours: Hardware-Enhanced Robust Optimized Post-Training Quantization Framework for W8A8 Transformers
}

\author{
Zhewei Yao, Reza Yazdani Aminabadi,  Stephen Youn, \\
Xiaoxia Wu, Elton Zheng, Yuxiong He
\\ Microsoft 
% \\ 
% {\tt \small\{\}@microsoft.com}
}

\date{}
\maketitle

%%%%%%%% BODY TEXT
\begin{abstract}
 Quantization techniques are pivotal in reducing the memory and computational demands of deep neural network inference. Existing solutions, such as ZeroQuant, offer dynamic quantization for models like BERT and GPT but overlook crucial memory-bounded operators and the complexities of per-token quantization. Addressing these gaps, we present a novel, fully hardware-enhanced robust optimized post-training W8A8 quantization framework, \ours. This framework uniquely integrates both memory bandwidth and compute-intensive operators, aiming for optimal hardware performance. Additionally, it offers flexibility by allowing specific INT8 modules to switch to FP16/BF16 mode, enhancing accuracy.
\end{abstract}
\section{Introduction}
\label{sec:intro}
Quantization is one of the most commonly used techniques to reduce the memory footprint and compute cost for deep neural network inference. 
Various quantization methods~\cite{dong2019hawq,gholami2021survey} have been proposed to speedup the inference and/or improve the throughput. 

There are mainly two approaches to realize quantization for a trained model:
\begin{enumerate}
    \item \textbf{Quantization aware training (\qat).} \qat~\cite{shen2020q,zafrir2019q8bert,fan2020training,zhang2020ternarybert,bai2020binarybert,hinton2015distilling,jiao2019tinybert,jin2021kdlsq,esser2019learned,tao2022compression,yao2020hawqv3,dong2019hawqv2} generally leads to high-quality model but associated with high training/finetuning cost.
    \item \textbf{Post-training quantization (\ptq).} \ptq~\cite{zadeh2020gobo,bondarenko2021understanding,yao2022zeroquant,wu2023zeroquant,xiao2023smoothquant,yao2023zeroquant,wei2022outlier} minimizes the finetuning cost of \qat but lowers the model quality as compared to \qat.
\end{enumerate}
In practice, particularly for fast evolving domains, e.g., Ads and recommendation system, \ptq is preferred due to its lower cost and fast adoption speed. 
In order to alleviate the accuracy issue of \ptq, various methods have been proposed. 
However, due to the interdisciplinary gap between machine-learning algorithms and hardware (in this work, we mainly target Nvidia GPUs, e.g., A100), a hardware-aware \ptq method is still largely missing in this field, particularly for Transformer-based models. 

For instance, ZeroQuant~\cite{yao2022zeroquant} proposes dynamic per-token activation quantization and per-column weight quantization for \bert~\cite{devlin2018bert} and \gpt~\cite{radford2019gpt} models to achieve good accuracy. 
However, it does not consider (1) the non-trivial memory bounded operators, e.g., LayerNorm and attention, and leaves these parts in FP16/BF16 and (2) the per-token quantization cost of invoking additional kernel, when there is no fusion opportunity, e.g., the INT8 GeMM operator of the attention output linear layer. 

To resolve those limitations, we introduce \ours, a fully hardware-aware and practical post-training W8A8 quantization framework. 
Our contributions are summarized as below.
\begin{enumerate}
    \item \ours considers both memory bandwidth bound and compute intense operators into design. 
    As such, the framework can (potentially) achieve the best hardware performance.   
    \item To further improve the usability of \ours, different quantization levels, i.e., the ratio of INT8 operators vs. FP16/BF16 counterparts, of \ours can be performed to achieve desired accuracy and latency trade-off.
\end{enumerate}
\section{Methodology}
\label{sec:method}
\subsection{Quantization Schemes}
Throughout the work, we use symmetric uniform INT8 quantization unless specific comment is applied. 
However, our method also works for other 8-bit precision formats, like FP8. 
Particularly, we use the following column-major weight matrix format to perform GeMM, 
\begin{equation}
    Y = XW,
\end{equation}
where $X\in\R^{n\times d}$ is the activation, and $W\in\R^{d\times m}$ is the weight matrix. 
For weight quantization, we perform column-wise quantization~\cite{yao2022zeroquant}, i.e., each column of the weight has its own scaling factor,
\begin{equation}
    W = W_{int8}S_w,
\end{equation}
where $W$ is the reconstructed weight matrix, $W_{int8}$ is the INT8 counterpart, and $S_w\in\R^{1\times m}$ is the scaling vector.\footnote{Note that here we use PyTorch/Numpy friendly calculation, i.e., $W_{int8}S_w=W_{int8}\text{Diag}(S_w)$, where $\text{Diag}$ is used to make the vector as the diagonal of the matrix.}
For activation quantization, we apply three different quantization schemes and we will explain in the next section for the utilization of them.

\noindent\textbf{Token-wise quantization (\twq)} The first quantization scheme we use for token quantization is \twq~\cite{yao2022zeroquant}, i.e., 
\begin{equation}
X = S_xX_{int8},
\end{equation}
where $X$ is the reconstructed activation, $X_{int8}$ is the INT8 counterpart, and $S_x\in\R^{n\times 1}$ is the scaling vector.
This approach requires the scaling vector $S_x$ to be calculated on-the-fly, which is more suitable to be fused with bandwidth bounded operators, like Layer Normalization (LN). In fact, quantization is done at zero memory-overhead cost, using extra register-level operations to compute min and max, to reduce the LN's output precision which is going to be used in the following GeMM operation. On the other hand, this approach of scaling hurts Tensor-core efficiency if fused with compute-bound operations such as GeMMs, due to increasing the register pressure and add more compute per Matrix-Multiply-Accumulate (MMA) operations.

\noindent\textbf{Feature-wise quantization (\fwq)} The second quantization scheme we use for token quantization is \fwq~\cite{wei2022outlier,xiao2023smoothquant}, i.e., 
\begin{equation}
X = X_{int8}S_x,
\end{equation}
where $S_x\in\R^{1\times d}$ is the scaling vector. $S_x$ here needs to be calibrated in the pre-processing phase, i.e., feeding multiple batches of data in the network to get the scaling factor. 
As it is pre-determined, it can be simply fused with most other operators. Compared to TWQ quantization scheme, that involves reading a token-width length to quantize data and can be only fused with certain operations, FWQ scaling can be fused with either memory-bound or compute-bound operations. 

\noindent\textbf{Static quantization (\sq)}
The final approach we use here is \sq~\cite{gholami2021survey}, i.e., 
\begin{equation}
X = X_{int8}S_x=S_xX_{int8},
\end{equation}
where $S_x\in\R$ is just a single real value. Similar to \fwq, it also needs to be calibrated in the pre-processing phase.  

\subsection{Core methodology}
We discuss the three main components of \ours in this section.
\subsubsection{Embedding Quantization}
\label{sec:embedding_module}
The first main operator of Transformer models is the lookup table, aka embedding. 
Normally, there are three types of embedding, i.e., token embedding ($X_t$), position embedding ($X_p$), and sentence type embedding ($X_s$). 
When the batch size is large enough, the latter two, i.e., $X_p$ and $X_s$ are relative small as compared to $X_t$. 
When we have all embedding, a layer norm is applied to get the final result, i.e.,
\begin{equation}
    X_{emb} = LN(X_t,~X_p,~X_s),
\end{equation}
where $X_{emb}$ is the output of the layer norm and $LN$ applies layer norm operator of the sum of all its inputs.
The $LN$ operator here is a memory bandwidth bounded operator related to the input $X_t$ and the output $X_{emb}$. 
In order to reduce the memory-bandwidth overhead, we perform \twq on both $X_t$ and $X_{emb}$, i.e., 
\begin{equation}
    S_{emb}X_{emb,int8} = LN^{quant}(S_{x_t}X_{t,int8},~X_p,~X_s),
\end{equation}
where $LN^{Quant}$ is a quantization-aware operator. 
By utilizing the above embedding format, we roughly reduce the data volume communicated for the following operation by 2x.

\subsubsection{Attention Module Quantization}
\label{sec:attn_module}
\begin{figure}
    \centering
     \includegraphics[width=1.02\textwidth]{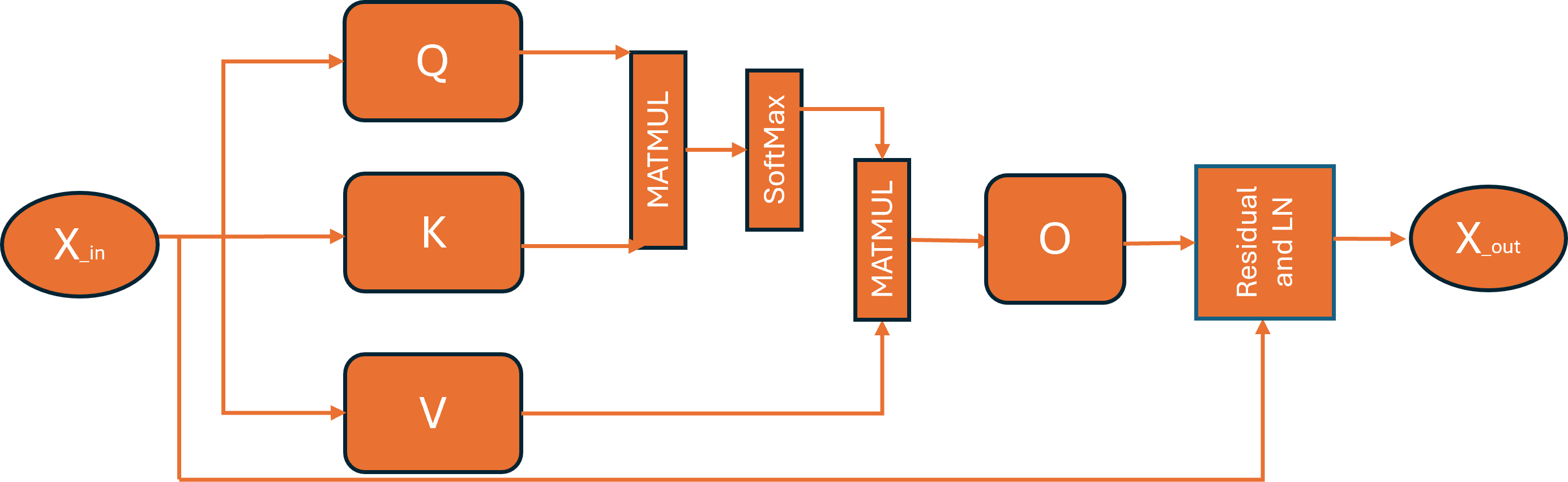}
    \caption{
The illustration of attention module in a transformer layer.}
\label{fig:attn_module}
\end{figure}
The attention module is illustrated in~\fref{fig:attn_module}. 
We give a high level calculation process of the attention module here and please refer to~\cite{vaswani2017attention} for more details, 
\begin{align}
    X_{q/k/v} &= X_{in}W_{q/k/v},\\
    A &= X_q X_k^T / \sqrt{d}, \\
    P &= Softmax(A), \\
    X_{attn} &= PX_v, \\
    X_o &= X_{attn}W_O,\\
    X_{out} &= LN(X_{in},~X_o). 
\end{align}
Before diving into more details, we first categorize all activation quantization schemes:
\begin{itemize}
    \item \twq is applied for $X_{in}$ and $X_{out}$ to preserve high accuracy for the input and output of a transformer layer with the least performance overhead since the scaling logic can be fused in the LN operations happening beforehand.
    \item \sq is applied for $X_q$, $X_k$, $X_v$, and $P$, in order to improve the efficiency of the GeMM operations that involves these tensors. We have this logic added onto the flash-attention kernel implementation, and the dtype for each GeMM can be configured in order to preserve the model accuracy.
    \item \fwq is applied for $X_{attn}$ and $X_o$ to reduce the complexity of scaling activation for the GeMM operation while preserving the accuracy. Compared to \sq, we are using one scale per output element, so the performance cost of this operation is similar to adding a bias at the linear layer.
    \item For $A$, no quantization is applied. This is due to the sensitivity of the attention score values to their precision which could hurt the model accuracy on downstream tasks.
\end{itemize}
Before applying weight quantization, we have
\begin{align}
    X_{q/k/v,int8}S_{q/k/v} &= S_{in}X_{in,int8}W_{q/k/v},\\
    A &= S_qS_kX_{q,int8} X_{k,int8}^T / \sqrt{d}, \\
    S_{p}P_{int8} &= Softmax^{Quant}(A), \\
    X_{attn, int8}S_{attn} &= S_pS_vP_{int8}X_{v,~int8}, \\
    X_{o, int8}S_o &= X_{attn, int8}S_{attn}W_O,\\
    S_{out}X_{out, int8} &= LN^{Quant}(S_{in}X_{in, int8},~X_{o, int8}S_{o}). 
\end{align}
Here, $\cdot^{quant}$ is the quantization-aware operator, and the output $Softmax^{quant}$ is asymmetric INT8 numbers since there is no negative value in the output of softmax. 
Now, let's dive deeper into weight quantization and the GeMM operator. 
First of all, we can apply the same kernel fusion to fuse the dequantization operator with INT8 GeMM as~\cite{yao2022zeroquant}. 
To further reduce the quantization overhead, we could fuse the scaling factors of \fwq and \sq into the INT8 GeMM as the scaling factors are pre-determined without any on-the-fly reduction operator.\footnote{Please refer to \url{https://github.com/openai/triton/blob/main/python/tutorials/03-matrix-multiplication.py} as an example of post-GeMM operator fusion.}
More importantly, the \fwq/\sq quantization can be simplified as an simple round-to-integer operator without any division/multiplication, as the scaling factor can be merged into weight matrix. 
Taking $X_{q, int8}$ as an example, we can define 
\begin{align}
    \tilde W_q &= W_q / S_q,\\
    \tilde W_{q,int8}S_q &= Quant(\tilde W_q).
\end{align}
Here $Quant$ is the quantization converting operator.
Afterwards, the post-GeMM quantization operator is simplified as 
\begin{equation}
    X{q, int8} = Round\left( GeMM^{quant}(X_{in, int8},~W_{q, int8},~S_{in},~S_{q})\right),
\end{equation}
where $Round(\cdot)$ is the round to the integer operator. 
Similarly, we do not need to dequantize the calculation of $A$ following by the division by $\sqrt{d}$.
We could simplify it with $\tilde d = S_qS_k/\sqrt{d}$ and $A = GeMM^{quant}(X_{q,int8}, X_{k,int8}^T, \tilde d)$. 

The scaling factor of both $S_{attn}$ and $S_{o}$ can be merged into $W_o$ by 
\begin{equation}
    \tilde W_o = S_{attn}W_o / S_o.
\end{equation}
Such that the overall kernel implementation can be significantly simplified. 
Afterwards, the $LN^{quant}$ operator takes two INT8 number as input and outputs the final INT8 activation for the following MLP module.\footnote{We use $\cdot^{quant}$ operator in a unified way even though the inputs (e.g., the number of input variables, the data type) and/or the outputs are not the same, e.g., the $LN^{quant}$ used in~\sref{sec:embedding_module} and \sref{sec:attn_module} are two different kernels.}

\subsubsection{MLP Module Quantization}
\begin{figure}
    \centering
     \includegraphics[width=1.0\textwidth]{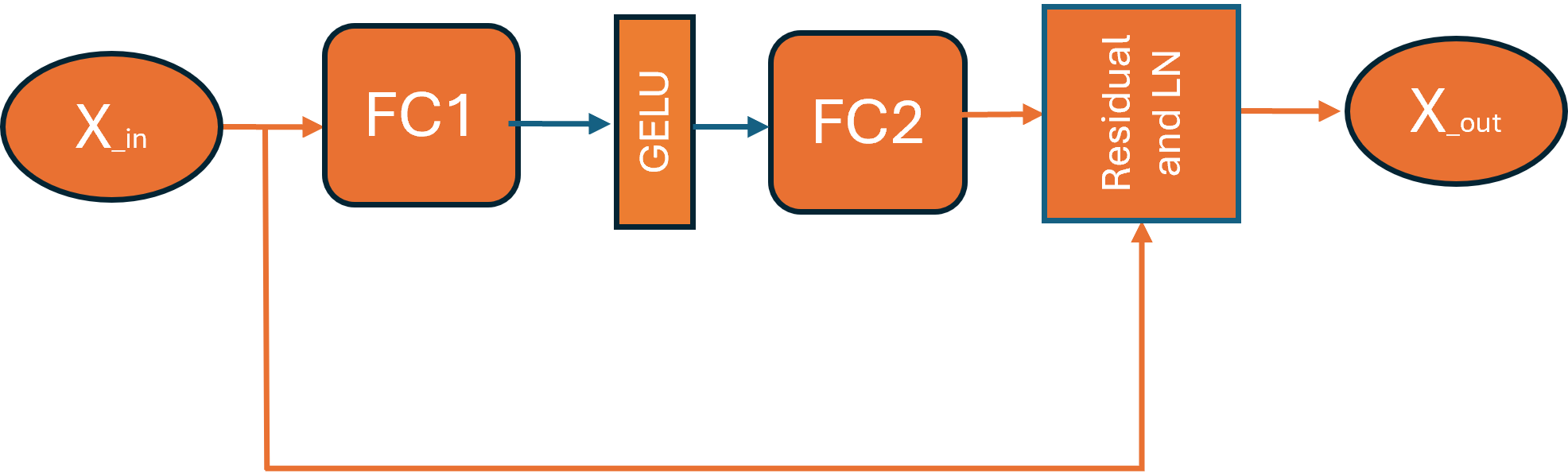}
    \caption{
The illustration of MLP module in a transformer layer.}
\label{fig:mlp}
\end{figure}
A standard MLP module is illustrated in~\fref{fig:mlp}. 
The mathematical flow of the module is as
\begin{align}
    X_1 &= X_{in}W_1,\\
    A &= GELU(X_1), \\
    X_2 &= AW_2, \\
    X_{out} &= LN(X_{in}, X_2). 
\end{align}
Similar as before, we first categorize all activation quantization schemes.
For $X_{in}$ and $X_{out}$, \twq is applied.
For $A$ and $X_2$, \fwq is applied. 
For $X_1$, no quantization is used. 
Before applying weight quantization, we have
\begin{align}
    X_1 &= S_{in}X_{in,~int8}W_1,\\
    A_{int8}S_a &= GELU^{quant}(X_1), \\
    X_{2, int8}S_{x_2} &= A_{int8}S_aW_2, \\
    X_{out} &= LN^{quant}(S_{in}X_{in},~X_2S_{x_2}). 
\end{align}
Similar as before, the scaling factors, i.e., $S_a$ and $S_{x_2}$, can be merged into $W_2$ to simplify calculation:
\begin{align}
    \tilde W_2 = S_aW_2 / S_{x_2}.
\end{align}

\subsection{Mixed Precision Inference}
Combining all techniques in the previous section, we get the final \ours design. 
However, different models and/or tasks have different tolerance to quantization, and they also have different desire on the trade-off of accuracy and system efficiency. 
In order to meet requirements for various models/tasks, mixed-precision inference is one of the solutions for quantization. 

Thanks to the modulized design of \ours, we can set various quantization level for our final model. 
To demonstrate the necessary of mixed-precision inference, we show the accuracy of three quantization levels (\tref{tab:mode}) in the next section.

\begin{table}[t]
\caption{
Different quantization mode of \ours. 
\checkmark ~means using INT8 and \xmark ~means using FP16/BF16.
}\centering
\label{tab:mode}
\begin{adjustbox}{width=0.999\linewidth}
\centering
\begin{tabular}{lcccccccccccccc}
\toprule
Mode & Embedding & QKV GeMM & Attn. & Attn. Output & FC1 & FC2 \\
\ours-M1 &\checkmark &\checkmark &\xmark &\xmark &\checkmark &\xmark \\
\ours-M2 &\checkmark &\checkmark &\checkmark &\checkmark &\checkmark &\xmark \\
\ours-M3 &\checkmark &\checkmark &\checkmark &\checkmark &\checkmark &\checkmark \\
\bottomrule
\end{tabular}
\end{adjustbox}
\end{table}

\section{Result}
\label{sec:result}
\paragraph{Experiments Setting.}
We use the ``yoshitomo-matsubara/bert-base-uncased-'' family models from Huggingface~\cite{wolf2019huggingface} to test the accuracy of \ours. 
Particularly, we use 100 batches and batch size 16 to calibrate (i.e., only run the forward pass) all quantization-related values. 
The sequence length is 128 for all tasks.

\paragraph{Results.}

\begin{table}[ht]
\caption{Results of \ours for $\bert_{base}$ on GLUE benchmark (validation).}\label{tab:result}
\begin{adjustbox}{width=0.99\linewidth}
\centering
\begin{tabular}{lccccccccccccccc }
\toprule
Mode   &  CoLA  & MNLI-m/-mm  &  MRPC      &  QNLI  &    QQP      & RTE    & SST-2 &   STS-B      \\
  & Mcc     &   Acc/Acc   & F1/Acc     &  Acc   &  F1/Acc     & Acc    &  Acc  & Pear/Spea    \\   \midrule  
FP16 &61.05 &84.20/84.67 &90.68/87.25 & 91.58 & 87.83/90.95 & 67.51 &92.54 &88.88/88.55\\
\ours-M1 & 60.39 &84.29/84.52 &90.11/86.27 &91.51 &87.85/90.96 &68.59 &92.78 &88.78/88.47\\
\ours-M2 & 59.47 &84.06/84.67 &90.62/87.01 &91.51 &87.83/90.94 &67.51 &92.55 &88.86/88.51\\
\ours-M3 & 41.65 &83.61/84.17 &89.48/85.54 &91.31 &87.51/90.55 & 69.31 &92.20 &88.90/88.47\\
 \bottomrule
\end{tabular}
\end{adjustbox}
\end{table}

The results of different quantizaton level of \ours are shown in~\tref{tab:result}. 
The overall accuracy degrades as we increase the quantization level. 
However, besides CoLA, which is a super sentitive task, for all rest tasks, even \ours-M3 achieves reasonable accuracy drop as compared to FP16 counterpart. 

\paragraph{Discussion.}
Note that the main focus of this work is not to achieve the best accuracy but to show the hardware-aware and practical INT8 PTQ framework, \ours. 
As such, we did not tune any hyperparameters, including both explicit  and/or implicit hyperparameters.
For instance,
(1) for explicit hyperparameters, we did not change calibration iterations. 
By reducing the batch number from 100 to 5 for CoLA, \ours-M3 can get about 1\% gain as compared to the result reported in~\tref{tab:result};
(2) for implicit hyperparameters, we did not tune the min/max value truncation for quantization. 
Normally, a careful tuned quantization threshold can boost the accuracy~\cite{wu2023understanding}.

Two big missing pieces of the current work are the kernel implementation and the end-to-end system performance measurement. 
We leave them as a future work.
\section{Conclusion}
\label{sec:conclusions}
In this study, we explored the intricacies of quantization for optimizing inference of transformer-based models, with a spotlight on Post-training Quantization (PTQ). Addressing the challenge of aligning algorithms with hardware, we introduced \ours, a novel hardware-enhanced post-training W8A8 quantization framework. Our experiments, based on the Huggingface model family, demonstrated the efficacy of \ours, highlighting its potential even with increased quantization levels. Areas like kernel implementation and end-to-end system performance measurement remain unexplored, paving the way for future research.

{
\bibliographystyle{plain}
\bibliography{ref.bib}
}

\end{document}